# AN EXPLANATION MECHANISM FOR BAYESIAN INFERENCING SYSTEMS

Steven W. Norton[*]
PAR Government Systems Corporation
New Hartford, NY 13413

Abstract: Explanation facilities are a particularly important feature of expert system frameworks. It is an area in which traditional rule-based expert system frameworks have had mixed results. While explanations about control are well handled, facilities are needed for generating better explanations concerning knowledge base content. This paper approaches the explanation problem by examining the effect an event has on a variable of interest within a symmetric Bayesian inferencing system. We argue that any effect measure operating in this context must satisfy certain properties. Such a measure is proposed. It forms the basis for an explanation facility which allows the user of the Generalized Bayesian Inferencing System to question the meaning of the knowledge base. That facility is described in detail.

## 1. Introduction

The area of Expert Systems (ES) is currently one of considerable interest. Much of this interest arises because ES technology is returning real benefits to ES users [1]. It is fair to say, however, that ES technology has not yet fully matured. A complete, fully functional ES framework must provide a range of features. One of these is an explanation capability. Explanation is particularly important since in most critical applications the human user bears the ultimate responsibility for action. In medicine, for example, the human practitioner utilizing an ES as a consultant requires an explanation of the machine inferencing process whenever its recommendation is not precisely as expected. There are many other critical areas in which expert systems will be called upon to explain themselves (nuclear, military, etc.).

The explanation facilities of current ES frameworks are chiefly directed at control mechanisms [3,8]. However, we often desire explanations of the contents of the knowledge base itself. A summary of how particular values in the knowledge base were computed may be insufficient. The uncertainty representations and inferencing mechanisms of various rule based expert systems have impeded the development of robust explanation capabilites through ill-defined actions and inconsistent semantics.

This paper addresses both theoretical and practical issues related to explanation. In Bayesian inferencing systems, not only can explanations of control be generated, but also meaningful explanations of database content. We propose a measure of the effect that an event has on a variable of interest. That measure, summarizing correlation information, prior probabilities, and posterior probabilities, is at the core of the explanation facilities of the Generalized Bayesian Inferencing System (GBI). In the following sections we detail the most relevant features of GBI, of our effect measure, and of our explanation facility. Finally, we describe a number of enhancements possible within our implementation which would significantly enhance its explanation capabilities.

## 2. The Generalized Bayesian Inferencing System

The Generalized Bayesian Inferencing System is a framework for building expert systems, which supports inferencing under uncertainty according to the Bayesian hierarchical inferencing paradigm. It differs significantly from other systems reasoning under uncertainty (PROSPECTOR, MYCIN) in several respects, and the reader is referred to [2,5,6,7] for more detail. In this secition, we highlight the features of GBI most relevant to the topic of explanation and provide an example to fuel further discussion.

---

[*] Now employed by Siemens Research and Technology Laboratories, 105 College Road East, Princeton, NJ 08540. Phone: (609) 734-6500 .

193

GBI's knowledge base is a network of probability distributions over intersecting sets of events. Each event set is a Local Event Group (LEG [4]). It should be a set of importantly correlated variables. Since a LEG does not encompass all the problem variables, it has a marginal probability distribution called a Component Marginal Distribution (CMD [5]). A LEG Network (LEG Net) is made up of several LEGs, some of which share common variables. Two LEGs sharing common variables must be consistent, meaning that the computation of joint probabilities over those common variables yields identical results no matter which CMD is used.

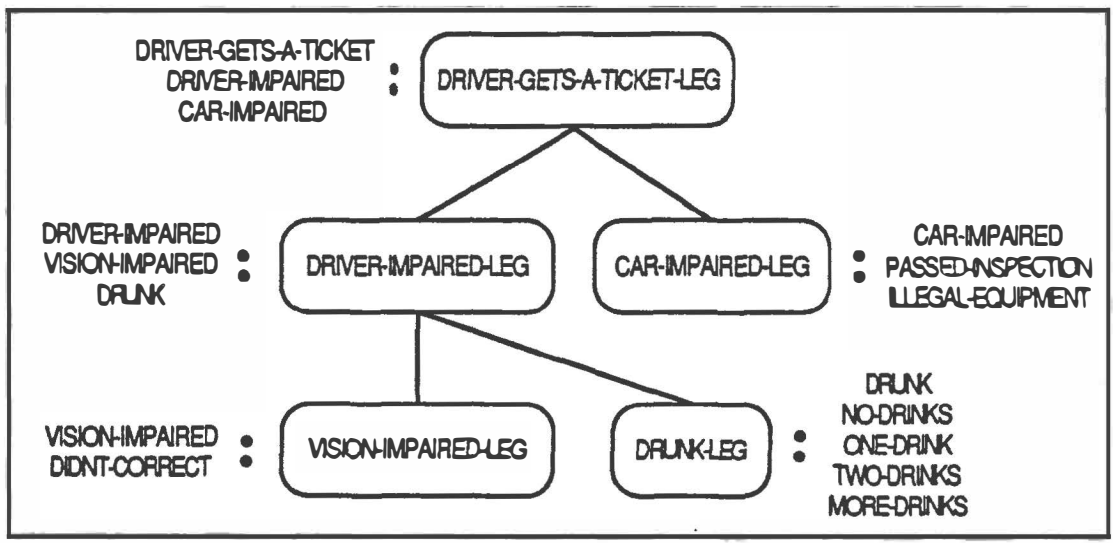

Figure 1: A Sample LEG Network

Figure 1 contains a simple example network for reasoning about whether or not the driver of an automobile receives a ticket at the scene of an accident. The events are listed beside each LEG. It's an obvious simplification since it doesn't consider any parameters of the accident itself. If the observation is made that TWO-DRINKS occured, the CMD associated with DRUNK-LEG is altered. Using the updating rule given in [2], the effect of the observation is propagated to the DRIVER-IMPAIRED-LEG through the action of the event DRUNK. If the CMD for the DRIVER-IMPAIRED-LEG is actually changed, then that effect will have to be propagated to the VISION-IMPAIRED-LEG and to the DRIVER-GETS-A-TICKET-LEG. If the CMD for DRIVER-GETS-A-TICKET-LEG changes, then the CAR-IMPAIRED-LEG will be updated as well.

### 3. Explanation Facilities

At some point during a consultation session with GBI, the user will want an explanation of the expert system's behavior. In this paper, we address neither the natural language aspects of automatically generated explanations nor the explanation of control decisions. These topics have been covered elsewhere in the literature for other expert system frameworks. The same techniques may be used here. Instead we focus on explaining the contents of the LEG Net, and why changes occured in the LEG Net in order to highlight the possibilities for explanation in Bayesian expert systems.

The GBI framework was created out of a desire to incorporate correlations between variables directly into the knowledge base of an expert system. In fact, the correlations guide LEG Net construction and CMD specification [7]. Two events are correlated if information about one yields information about the other. For example, the probability of DRIVER-GETS-A-TICKET given CAR-IMPAIRED is greater than the probability of DRIVER-GETS-A-TICKET. Hence, those events are positively correlated. CAR- IMPAIRED and PASSED-INSPECTION are negatively correlated since the probability of CAR-IMPAIRED given PASSED-INSPECTION is less than the probability of CAR- IMPAIRED. This can serve as a basis for generating explanations. Why did an event become more likely? Either because a positively correlated event became more likely, or because a negatively correlated event became less likely.

We require a class of explanations for the GBI framework generated solely within a single LEG, after a single evidence update. We classify any explanation which makes explicit use of information about the exact sequence of evidence observations in order to trace back from a hypothesis to that particular evidence variable as an explanation of control information. While this type of explanation is important, we leave it out of the present discussion. The decision is not entirely arbitrary since all the available evidence could be gathered and its effects propogated in a single step using the GBI framework. In that case, there is no sequencing of evidence observations. Even in the event that evidence updates occur successively, there may be multiple paths through the LEG Net leading from the evidence to the hypothesis of interest. But most of all, there may be no direct relationship (in terms of correlations) within a given LEG between the variable which is changing (or any joint variable containing it) and the hypothesis of interest. Such examples are not hard to develop. Let a local explanation be any explanation for a change in a variable of interest which is expressed solely in terms of variables coexisting with it in a LEG.

Suppose that three updates altered the DRIVER-GETS-A-TICKET network, that the first was from CAR-IMPAIRED-LEG, the second from DRUNK-LEG, and the third from VISION-IMPAIRED-LEG. During the first update, PASSED-INSPECTION occurred, and ILLEGAL-EQUIPMENT did not. During the second update, NO-DRINKS occurred. Focusing our attention after the second update and asking GBI for a local explanation of DRIVER-GETS-A-TICKET might yield the explanations given in Figure 2. The first one was generated with a level of detail appropriate for a typical user. In the second example the level of detail is more appropriate for a knowledge engineer.

---

The probability of DRIVER-GETS-A-TICKET decreased because the probability of DRIVER-IMPAIRED decreased after the update of DRUNK-LEG.

Events DRIVER-GETS-A-TICKET and DRIVER-IMPAIRED are positively correlated ( P[ DRIVER-GETS-A-TICKET | DRIVER-IMPAIRED ] - P[ DRIVER-GETS-A-TICKET] = 0.70 ) . The probability of DRIVER-GETS-A-TICKET decreased ( from 0.09 to 0.06 ) because the probability of DRIVER-IMPAIRED decreased ( from 0.05 to 0.01 ) after the update of DRUNK-LEG.

---

Figure 2: Local Explanations

In order to generate this type of explanation, we are proposing a heuristic measure of the impact an event has had on a variable of interest within the same LEG. To quantify this effect (here the effect of some evidence on a hypothesis variable), a three step calculation is performed. The first part is a measure of correlation equal to the difference between the probability of the hypothesis and the evidence, and the product of their marginal probabilities. The magnitude of the effect must increase with the amount of change in the probability of the either event, and so the second and third parts of the calculation are the amounts by which the probabilities of the hypothesis and evidence change. We are left with a quantity which can always be maximized to achieve an appropriate result. Thus, if we are interested in the effect an evidence event $E$ has had on a hypothesis variable $H$, we compute it as in Equation 1.

$$Ef(H,E) = [\ Pr(HE) - Pr(H)Pr(E))\ ] \times [\ Pr\,'(H) - Pr(H)\ ] \times [\ Pr\,'(E) - Pr(E)\ ] \qquad (1)$$

This measure, the Effect, is intuitively pleasing. Compare two different Effects, $Ef_1$ and $Ef_2$, computed for a single hypothesis variable. They can be expressed as the products of correlations, changes in the evidence probabilities, and the change in the hypothesis probability. Let these be $Corr_1$, $Delta_1$, $Corr_2$, $Delta_2$, and $Delta_h$ so that $Ef_1 = Corr_1 \times Delta_h \times Delta_1$ and $Ef_2 = Corr_2 \times Delta_h \times Delta_2$. Assume that $Delta_h$ is positive, indicating that the hypothesis variable became more likely. If the changes in the evidence probabilities are positive and equal, then the greater Effect will come from the variable with the larger correlation. If the changes in the evidence probabilities are negative and equal, then the greater Effect will come from the variable with the smaller correlation. And if the correlations are positive and equal,

195

the greater Effect will come from the evidence variable which undergoes the greater positive change. The same analysis shows the measure is reasonable when the Delta$_h$ term is negative.

There are two properties that an Effect measure should have, stemming from desirable properties of Bayesian systems. The first is symmetry: Ef(H,E) = Ef(E,H). Symmetry is desirable because in a symmetric Bayesian system (such as GBI) initiating an update with a change in evidence has similar (if not identical) consequences to initiating an update with a change in the hypothesis. The other property is that Ef(H, not E) = Ef(H,E). Since a Bayesian system requires that the probability of an event occurring and the probability of the same event not occurring sum to one (in the limit anyway), we cannot distinguish between the effect of those opposites except by introducing additional structure.

| GBI Explanation Window | Causal<br>Diagnostic | **Local**<br>Global |
|---|---|---|
| | User<br>**Knowledge Engineer** | **Use-Current-Data**<br>Use-All-History |
| Local Event Groups (LEGs)<br>**DRIVER-GETS-A-TICKET-LEG**<br>DRIVER-IMPAIRED-LEG<br>CAR-IMPAIRED-LEG | | |
| Events Within the Highlighted LEG<br>**DRIVER-GETS-A-TICKET**<br>DRIVER-IMPAIRED<br>CAR-IMPAIRED | | |
| Events DRIVER-GETS-A-TICKET and DRIVER-IMPAIRED are positively correlated ( P[ DRIVER-GETS-A-TICKET \| DRIVER-IMPAIRED ] - P[ DRIVER-GETS-A-TICKET ] = 0.70 ). The probability of DRIVER-GETS-A-TICKET decreased ( from 0.09 to 0.06 ) because the probability of DRIVER-IMPAIRED decreased ( from 0.05 to 0.01 ) after the update of DRUNK-LEG.<br><br>Explanation Typeout Window | | |
| **Explain**   When   Clear   Initialize   Structure   Help | | |

Figure 3: The GBI Explanation Facility

The measure in Equation 1 satisfies both of the above properties. It is the basis for the current GBI explanation capability. We have only implemented explanation for the univariate case, but have plans to extend it since the measure can clearly be applied to generation of multivariate explanations. For example, the joint event DRIVER-IMPAIRED & CAR-IMPAIRED may have a stronger effect on DRIVER-GETS-A-TICKET than either evidence variable alone. There may also be cases in which we want a change in a joint event to be explained.

An alternative measure of effect is the liklihood ratio used in Prospector [4]. We note, however, that it is a measure of potential effect rather than actual effect. As such, it is most useful (directly so) in controlling inference and subsequently explaining control decisions. If the liklihood ratio is substituted for the correlation measure in Equation 1, the measure fails because the liklihood ratio is non-negative. Using one minus the liklihood ratio also fails since the liklihood ratio ranges between zero and infinity. The result is to favor events positively correlated with the hypothesis over events negatively correlated with the hypothesis.

Figure 3 portrays the GBI explanation facility in action. In the upper righthand corner are switches which control the explanation. Below the switches are the LEGs and Events menus, as well as the explanation typeout window and the explanation command menu. The LEGs menu allows the user to roam within the LEG Net. As he does so, the Events menu reflects the contents of the current LEG. In the figure, LOCAL and KNOWLEDGE ENGINEER switches are set to generate a detailed local explanation.

Since several updates occur during a typical session, the GBI explanation facility needs mechanisms to handle sequenced updates. While the fact that updates occur sequentially is used explicitly in the explanation computations, information about the particular evidence variables observed is not.[1] The WHEN option from the command menu allows the user to focus on a particular evidence update. In that context, the USE-CURRENT-DATA and USE-ALL-HISTORY switches determine the temporal extent of the explanation. If USE-CURRENT-DATA is selected, the current update alone is used to generate the explanation. On the other hand, if USE-ALL-HISTORY is selected GBI summarizes the effects of all the earlier updates on the variable of interest. Figure 4 illustrates local explanations using a series of several updates. Given identical switch settings, the current GBI explanation facility presents this same information although with somewhat less polished delivery.

A limitation of this approach becomes apparent when, after the update of CAR-IMPAIRED-LEG, we ask for an explanation of CAR-IMPAIRED without moving to the CAR-IMPAIRED-LEG. Since the default mode of the GBI explanation facility is to look in the current LEG for an explanation, GBI may suggest that CAR-IMPAIRED became more likely because DRIVER-GETS-A-TICKET became more likely. If the evidence came from above rather than from below this would have been a pleasing explanation.

---

The probability of DRIVER-GETS-A-TICKET increased because the probability of CAR-IMPAIRED increased after the update of the CAR-IMPAIRED-LEG, and because the probability of DRIVER-IMPAIRED increased after the update of the DRUNK-LEG.

DRIVER-GETS-A-TICKET is positively correlated with CAR-IMPAIRED ( 0.73 ) and with DRIVER-IMPAIRED ( 0.80 ). The probability of DRIVER-GETS-A-TICKET increased ( from 0.05 to 0.35 ) because the probability of CAR-IMPAIRED increased ( from 0.05 to 0.60 ) after the update of the CAR-IMPAIRED-LEG, and because the probability of DRIVER-IMPAIRED increased ( from 0.02 to 0.40 ) after the update of the DRUNK-LEG.

---

Figure 4: Historical Explanations

---

[1] In the update to the LEG Net of Figure 1 due to the CAR-IMPAIRED LEG, two evidence variables are observed sumultaneously. To GBI, there is no sequencing of ILLEGAL-EQUIPMENT and PASSED-INSPECTION. Therefore a typical explanation which traces from DRIVER-GETS-A-TICKET to these evidence variables could only report that one of them was the cause. A mechanism such as the one we propose offers a more satisfactory explanation in this case.

In this particular case, we are seeking an explanation in terms of the causes of CAR-IMPAIRED. However, GBI itself has no real knowledge of causal structure. With the STRUCTURE option from the explanation command menu, the knowledge engineer is able to input structural information. Figure 4 shows the causal structure for our simple LEG Net. The link from DRIVER-IMPAIRED to DRIVER-GETS-A-TICKET may be read as "DRIVER-IMPAIRED causes DRIVER-GETS-A-TICKET" or "DRIVER-GETS-A-TICKET is a symptom of DRIVER-IMPAIRED." Strict interpretation of these relationships, at least in the context of this example, may not be very productive. Instead we view these links as additional control information for the explanation process itself. When the CAUSAL switch is set, GBI will only generate explanations in terms of the designated causes of the hypothesis of interest. DIAGNOSTIC, on the other hand, leads to the generation of explanations in terms of the hypotheses symptoms. Very simply, if either switch is activated, an additional filter is applied to candidate explanations. If the filter is so restrictive that GBI can not satisfy the request for explanation, it informs the user of the problem.

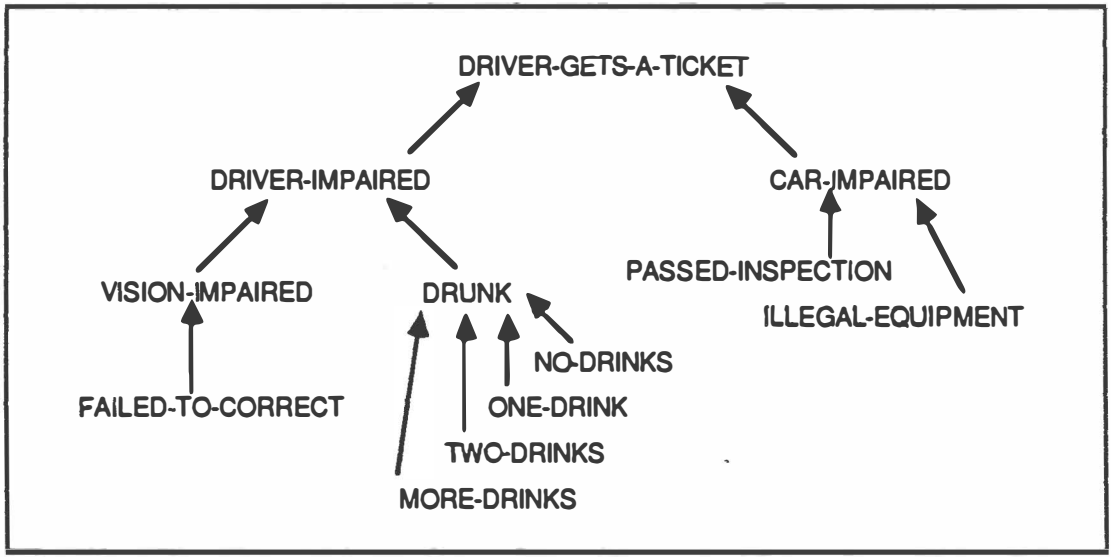

Figure 5: Causal Structure

The probability of DRIVER-GETS-A-TICKET increased because the probability of DRIVER-IMPAIRED increased, because the probability of DRUNK increased, because MORE-DRINKS occurred after the update of DRUNK-LEG.

DRIVER-GETS-A-TICKET is positively correlated ( 0.73 ) with DRIVER-IMPAIRED, which is positively correlated ( 0.80 ) with DRUNK, which is positively correlated ( 0.85 ) with MORE-DRINKS. The probability of DRIVER-GETS-A-TICKET increased ( from 0.05 to 0.80 ) because the probability of DRIVER-IMPAIRED increased ( from 0.04 to 0.65 ), because the probability of DRUNK increased ( from 0.01 to 0.25 ), because the probability of MORE-DRINKS increased ( from 0.10 to 1.00 ) after the update of DRUNK-LEG.

Figure 6: Global Explanations

The next logical step, now that the causal structure has been introduced, is to attempt to generate global explanations. For GBI this means tracing along the causal links looking at the most significant effects at each stage. Consequently, when the GLOBAL setting is in effect, the GBI explanation facility requires that either CAUSAL or DIAGNOSTIC be set. Figure 6 illustrates the use of the GLOBAL and CAUSAL switches. GBI delivers similar explanations when the GLOBAL and DIAGNOSTIC switches are set.

## 4. Concluding Remarks

The explanation capability of the Generalized Bayesian Inferencing System can be extended in several interesting ways. Two compatible avenues are available for improving the presentation of explanations. The first is simply to add a better natural language capability to replace the tedious fill-in-the-blank style. Natural language could even be used effectively in querying the system. The advantage of GBI is that there is a solid mathematical basis for explanations. The "richness" of information in the CMDs can provide a solid basis for generating and quantifying linguistic statements of relationships between variables. A further enhancement might be the graphical display of the LEG Net and causal structures. Not only would this replace the menu driven interface with something more convenient, but a graphical display would make the explanations much easier to understand. It would also make it easier for the user to direct the system towards the desired explanation whenever it strays.

Candidate computational changes include the availability of multivariate queries and multivariate explanations. Multivariate events might be taken as indirect causes for a change in a hypothesis variable, although that view is not entirely applicable under GBI. It is easy to think of instances when a joint event exerts the strongest influence on a hypothesis of interest. Inside a complicated LEG, these events could be the equivalent of the global explanations. Our system should also be able to recognize and then handle special relationships between variables such as mutual exclusion and implication.

Finally, while this paper has left out the issue of explaining control information, knowledge of control must be available for explanation. It can be used as is, or it can augment the kinds of explanations described above. It would be helpful to tell the user not only which variable had the greatest direct effect upon the hypothesis of interest, but also through which variable that effect had propogated in the path from the evidence.

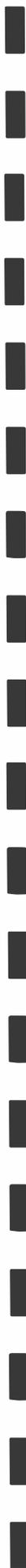